\documentclass{article}
\usepackage{spconf,amsmath,graphicx}
\usepackage{hyperref}
\usepackage{booktabs,threeparttable}
\usepackage{standalone}
\usepackage{multirow}
\usepackage{amssymb}
\usepackage{romannum}

\title{
Towards unsupervised speech recognition and synthesis\\
with quantized speech representation learning
}
\name{Alexander H. Liu$^\dagger$ \quad Tao Tu$^\dagger$ \quad Hung-yi Lee \quad Lin-shan Lee\thanks{ $^\dagger$ Indicates equal contribution.}}
\address{College of Electrical Engineering and Computer Science, National Taiwan University\\
         \small{\texttt{\{r07922013, r07922022,  hungyilee, lslee\}@ntu.edu.tw}
         }}
\begin{document}
\ninept
\maketitle
%

\newcommand{\QuatizeName}{Sequential Representation Quantization}

\begin{abstract}
In this paper we propose a Sequential Representation Quantization AutoEncoder (SeqRQ-AE) to learn from primarily unpaired audio data and produce sequences of representations very close to phoneme sequences of speech utterances.
This is achieved by proper temporal segmentation to make the representations phoneme-synchronized, and proper phonetic clustering to have total number of distinct representations close to the number of phonemes.
Mapping between the distinct representations and phonemes is learned from a small amount of annotated paired data.
Preliminary experiments on LJSpeech demonstrated the learned representations for vowels have relative locations in latent space in good parallel to that shown in the IPA vowel chart defined by linguistics experts.
With less than 20 minutes of annotated speech, our method outperformed existing methods on phoneme recognition and is able to synthesize intelligible speech that beats our baseline model.
\end{abstract}

\section{Introduction}
\label{sec:intro}

Speech signals are continuous in time, smoothly changing its diverse characteristics from time to time. 
Human listeners are able to divide the waveforms into small segments of variable lengths with relatively stable characteristics (temporal segmentation), and categorize the sounds of those segments into a finite number of recognizable clusters (phonetic clustering), producing linguistic units or phonemes, based on which humans learn to listen and speak started from infancy.
Training machines to perform the above temporal segmentation and phonetic clustering is not easy.
Hidden Markov Model performed on frame-level features were useful in early years, upon which various speech recognition~\cite{jelinek1976continuous} and synthesis~\cite{zen2009statistical} approaches were developed.
In the era of deep learning, learning representations for audio signals was considered as a promising approach, because temporal segmentation or phonetic clustering may be performed with these representations, or during the construction of theses representations.

So far, deep learning based speech representations were primary learned on frame level~\cite{oord2018representation,Schneider2019,chung2019unsupervised}.
However, without proper approaches to perform temporal segmentation and phonetic clustering, it is not easy to map such representations learned on frame level to linguistic units, and thus these representations cannot be reasonably interpreted by human.
Some higher level audio representations (e.g. audio word vectors) were also developed with the boundaries for the linguistic units needed as ground truth~\cite{chung2016audio} or automatically detected~\cite{wang2018segmental} but at the cost of the quality of the representations depended heavily on the accuracy of the boundaries.
In other words, temporal segmentation was the first gap to stride over, while phonetic clustering was the next.

Some recent works~\cite{chorowski2019unsupervised,van2017neural,baevski2019vq} successfully performed phonetic clustering to a good degree with proper quantization during learning the representations.
Nevertheless, the representations learned with frame-level quantization had a much higher diversity in acoustic characteristics not necessarily recognizable by human.
Also, without proper temporal segmentation, these learned representations are still far from desired human-like tasks such as automatic speech recognition (ASR) or text-to-speech (TTS).

In this paper, we seek to learn preliminary human-recognizable representations for speech signals from primarily unpaired audio data with a proposed framework Sequential Representation Quantization AutoEncoder (SeqRQ-AE).
With our proposed method, the learned sequence of representative vectors could be phoneme-synchronized (with proper temporal segmentation) and quantized into a number of clusters close to a number of phonemes (with proper phonetic clustering).
We used a small amount of paired data to map the quantized representations to a human-defined phoneme set, which allows us to interpret the learned representations and achieve initial speech recognition and synthesis tasks.

In our experiments, we demonstrated the learned representations for vowels have relative locations in latent space more or less parallel to that shown in the IPA vowel chart~\cite{international1999handbook} defined by human experts.
More importantly, the preliminary versions of ASR/TTS based on these learned representations are shown to perform better than existing similar approaches~\cite{tjandra2017listening,tjandra2019end,ren2019almost} based on human-defined phonemes, if only given a small amount of paired data.
These results verified that the learned representations are potentially useful in future human-like tasks such as ASR/TTS, especially with a very small amount of paired data or even unsupervised.

\begin{figure*}[ht]
\centerline{\includegraphics[width=18cm]{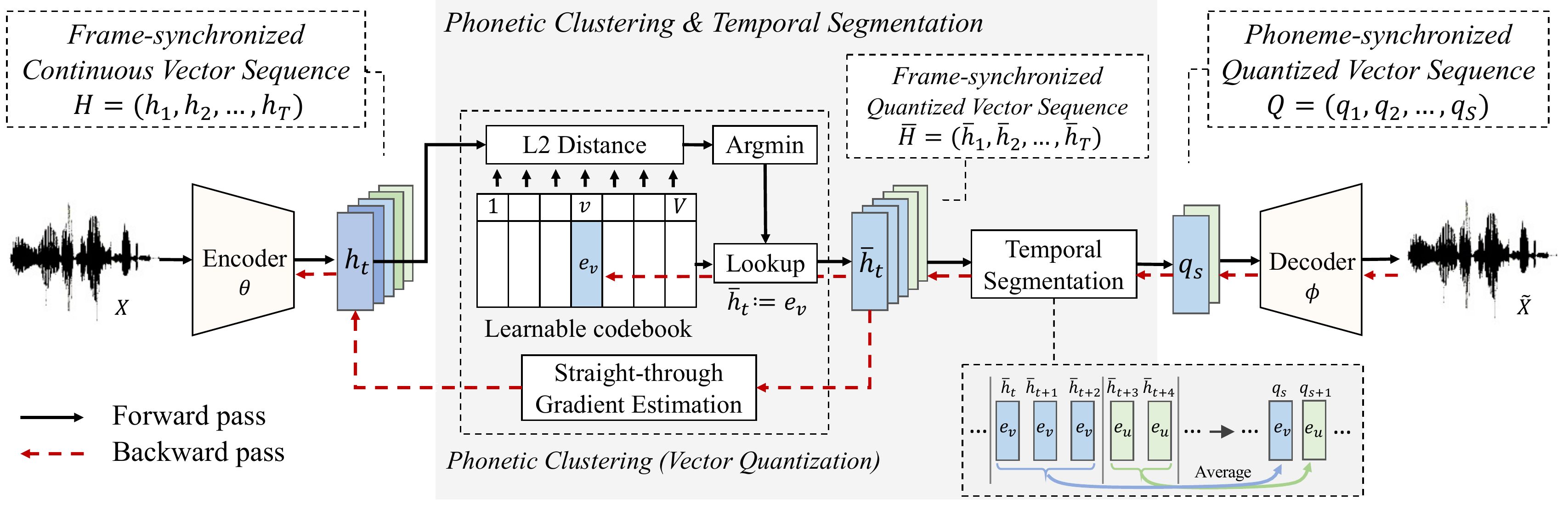}}

\caption{\small Overview of the proposed Sequential Representation Quantization AutoEncoder.
The input speech $X$ is first encoded into the frame.-synchronized continuous vector sequence $H$.
Next, Phonetic Clustering and Temporal Segmentation (see Sec.~\ref{subsec:vq}) is performed to obtain the phoneme-synchronized quantized vector sequence $Q$, which will be fed into a sequence-to-sequence decoder to reconstruct the input speech.
}
\vspace{-2pt}
\label{fig:overview}
\end{figure*}

\section{Proposed method}
\label{sec:method} 
Our goal is to learn a sequence of representations from the speech that matches the underlying linguistic unit sequence, which we use the phoneme sequence in this paper.
Fig.~\ref{fig:overview} gives an overview of our proposed framework and we organize our methodology as follows:
1) In Sec.~\ref{subsec:ae}, we first introduce a sequential auto-encoding framework that automatically learns to encode speech into a sequence of latent vectors that represents the input signal.
2) In Sec.~\ref{subsec:vq}, we perform Vector Quantization and Temporal Segmentation (shadowed part in Fig.~\ref{fig:overview}) to quantize similar vectors into the same codeword, and group consecutive same codewords into segments.
3) Finally, in Sec.~\ref{subsec:map}, we demonstrate how the quantized latent representation can be mapped to linguistic units with the aid of a limited amount of paired data.

\subsection{Representations with Sequential AutoEncoder}
\label{subsec:ae}

Given the input frame-level audio sequence $X = (x_1,x_2,...,x_T)$ with length $T$, an encoder network with parameter $\theta$ is employed to derive its corresponding sequence of latent representation 
\begin{equation}
    \label{eq:enc}
    H \equiv (h_1,h_2,...,h_T) = \text{Enc}_\theta(X),
\end{equation}
where $h_t \in \mathbb{R}^D$ for each time step $t$, and $D$ is the dimensionality of latent representation.
Since the representation sequence $H$ aligns to the input speech frames, we refer $H$ as a \textit{frame-synchronized} continuous vector sequence as shown in the left-hand side of Fig.~\ref{fig:overview}.

With the goal of learning a latent representation that is highly correlated to the linguistic units or human recognizable, we propose to perform \textit{Phonetic Clustering} and \textit{Temporal Segmentation} to simplify the frame-synchronized sequence $H$ into the \textit{phoneme-synchronized} quantized representation sequence $Q$, which is to be detailed in the next section.

To ensure the phoneme-synchronized sequence $Q$ is representative of input speech $X$, 
a sequence-to-sequence decoder is employed to reconstruct the input signal as follows:
\begin{equation}
    \label{eq:dec}
    \widetilde{X} = \text{Dec}_\phi(Q),
\end{equation}
where the sequence $\widetilde{X}$ is the frame-synchronized output of the decoder network with parameters $\phi$.

\vspace{-3pt}
\subsection{Phonetic Clustering \& Temporal Segmentation}
\label{subsec:vq}

This section includes vector quantization for phonetic clustering and the temporal segmentation to transduce the frame-synchronized representation sequence $H$ into the phoneme-synchronized sequence $Q$ as shown in the shadowed part in Fig.~\ref{fig:overview}.

\vspace{4pt}
\noindent \textbf{Phonetic Clustering.}
The input here is a sequence of continuous vectors $H$ in Eq.~(\ref{eq:enc}).
We borrow the discretization method for latent variables from Vector Quantised Variational AutoEncoder~\cite{van2017neural}.
To be more specific, we quantize each $h_t\in H$ to become an entry out of a learnable embedding table $E = \{e_1, e_2, ..., e_V\}$, which we refer to a \textit{codeword} $e_v$ in the \textit{codebook} $E$, with size $V$, and each $e_i \in \mathbb{R}^D$ as illustrated in Fig.~\ref{fig:overview}.

For each time step $t$, vector quantization is performed by replacing the encoder's output representation $h_t$ by its nearest neighbor (in terms of Euclidean distance) in the codebook.
Since selecting the closest entry (i.e. the Argmin operation in Fig.~\ref{fig:overview}) causes the quantization operation to be non-differentiable, the gradient of the encoder is approximated by straight-through (ST) gradient estimator~\cite{bengio2013estimating}. In practice, this can be addressed by having
\begin{equation}
    \label{eq:st}
    \bar{h}_t = h_t + e_{v} - \text{sg}(h_t), \text{~~~~~where~} {v} = \mathop{\arg\min}_{k} \| h_t - e_k\|_2
\end{equation}
and $\text{sg}(\cdot)$ is the stop-gradient operation that treats its input as constant during back-propagation.
Note that vector quantization is performing clustering with respect to the value of acoustic representation, we thus refer this operation \textit{Phonetic Clustering} in our proposed SeqRQ-AE.

\vspace{4pt}
\noindent \textbf{Temporal Segmentaion.} 
After phonetic clustering, the quantized sequence $\bar{H} = (\bar{h}_1,\bar{h}_2,...,\bar{h}_T )$ is still frame-synchronized.
To this end, we propose the temporal segmentation mechanism to produce the phoneme-synchronized quantized representation sequence $Q = (q_1,q_2,...,q_S)$ as illustrated in the lower right block of Fig.~\ref{fig:overview}.
This is done by simply grouping the consecutive repeated codewords within the sequence $(\bar{h}_1,\bar{h}_2,...,\bar{h}_T)$.

Temporal segmentation for continuous signals is not easy, but becomes easy after the vector quantization, because the input $\bar{H}$ here includes only $V$ distinct vectors.
Many vectors $h_t$ adjacent in time corresponding to signals with similar characteristics may be quantized to the same entry $e_v$ in the codebook.
So all we need to do is to group the consecutive repeated codewords $e_v$ into a segment.
Every change of the codeword, for example $e_v$ to $e_u$ at ${t+2}$ to ${t+3}$ in the lower right block of Fig.~\ref{fig:overview}, is a segment boundary.
Therefore each segment corresponds to a phonetic unit, and the output $Q$ is phoneme-synchronized.
Instead of discarding the repeated occurrence, we choose to take the average to stabilize the training of our proposed framework.

\vspace{-3pt}
\subsection{Quantized Representaion Mapping}
\label{subsec:map}

In the previous section, the quantized vectors in the codebook remain noninterpretable since reconstructing the input signal does not force the code in the codebook to be phoneme.
In this section, we demonstrate how each entry of the codebook can be mapped to a phoneme with a small amount of paired speech phoneme sequence data $(X_\text{pair},Y_\text{pair})$, where $X_\text{pair}=(x_1^\text{pair},x_2^\text{pair},...,x_T^\text{pair})$ is the frame-level audio sequence, and $Y_\text{pair}=(y_1^\text{pair},y_2^\text{pair},...,y_S^\text{pair})$ is the corresponding phoneme sequence.

We first set the size $V$ of the codebook  $E=\{e_1,e_2,...,e_V\}$ to be the number of all phonemes, and then assign each entry $e_v$ in $E$ to represent a phoneme $v$.
For each continuous representation vector $h_t$ from encoder, we define its probability of being mapped to a codeword $e_v$ in $E$ as
\begin{equation}
    \label{eq:prob}
    P(v|h_t) = \frac{\exp(- \| h_t - e_v\|_2)}{\sum_{k \in V} \exp(- \| h_t - e_k\|_2)},
\end{equation}
and the probability for some phoneme sequence $\tilde{Y}=(v_1,v_2,...,v_T)$ being the output from the encoder can be approximated by
\begin{equation}
    \label{eq:seq_prob}
    P(\tilde{Y}|H)  \approx  \prod_{t=1}^{T}{P(v_t|h_t)}.
\end{equation}

However, the above approximation requires the target sequence $\tilde{Y}$ to have length $T$ (i.e. $T$ frames).
But the phoneme-synchronized sequence $Y_\text{pair}$ has only $S$ phonemes, each may correspond to a number of repeated quantized codeword $e_v$.
This issue has been considered by connectionist temporal classification~\cite{graves2006connectionist} (CTC), so we have from CTC
\begin{equation}
    \label{eq:ctc_prob}
    P(Y_\text{pair}|H) = \sum\limits_{\tilde{Y} \in Y'}  P(\tilde{Y}|H),
\end{equation}
where $Y'$ is the set of all possible sequence $\tilde{Y}$ obtained by arbitrarily repeating elements of $Y_\text{pair}$ and/or inserting blank symbols until its length reaches $T$, the length of the encoder output sequence $H$. 
In other words, $Y'$ includes all possible $\tilde{Y}$ that reduces to $Y_\text{pair}$ via temporal segmentation.

For the decoder, the paired data can also be utilized given each entry of the codebook is matched to a phoneme.
We retrieve the embedding of each phoneme in $Y_\text{pair}$ from the codebook to obtain the ground truth phoneme embedding sequence $Q_\text{pair}$ and trained the decoder with standard sequence-to-sequence TTS objective~\cite{shen2018natural}.

The complete objective function of SeqRQ-AE can be written as
\begin{equation}
\begin{aligned}
    \label{eq:total}
    L_\text{total} = ~&\text{MSE}( \widetilde{X},X) \\
    &- \lambda_\text{1} \log P(Y_\text{pair}|H) \\
    &+ \lambda_\text{2}  \text{MSE}( \text{Dec}_\phi(Q_\text{pair}), X_\text{pair}) ,
\end{aligned}
\end{equation}
where the first term is the reconstruction loss of unpaired speech, the second term is the CTC loss from Eq.~(\ref{eq:ctc_prob}) for the phoneme sequence $Y_\text{pair}$ and the last term is the TTS loss for the target sequence $X_\text{pair}$.
We fix $\lambda_\text{1}$ and $\lambda_\text{2}$ to 0.5 throughout every experiment and train our proposed framework in an end-to-end manner without pre-training or fine-tuning.

\vspace{-11pt}

\begin{figure}[t]
\centerline{\includegraphics[width=8.5cm]{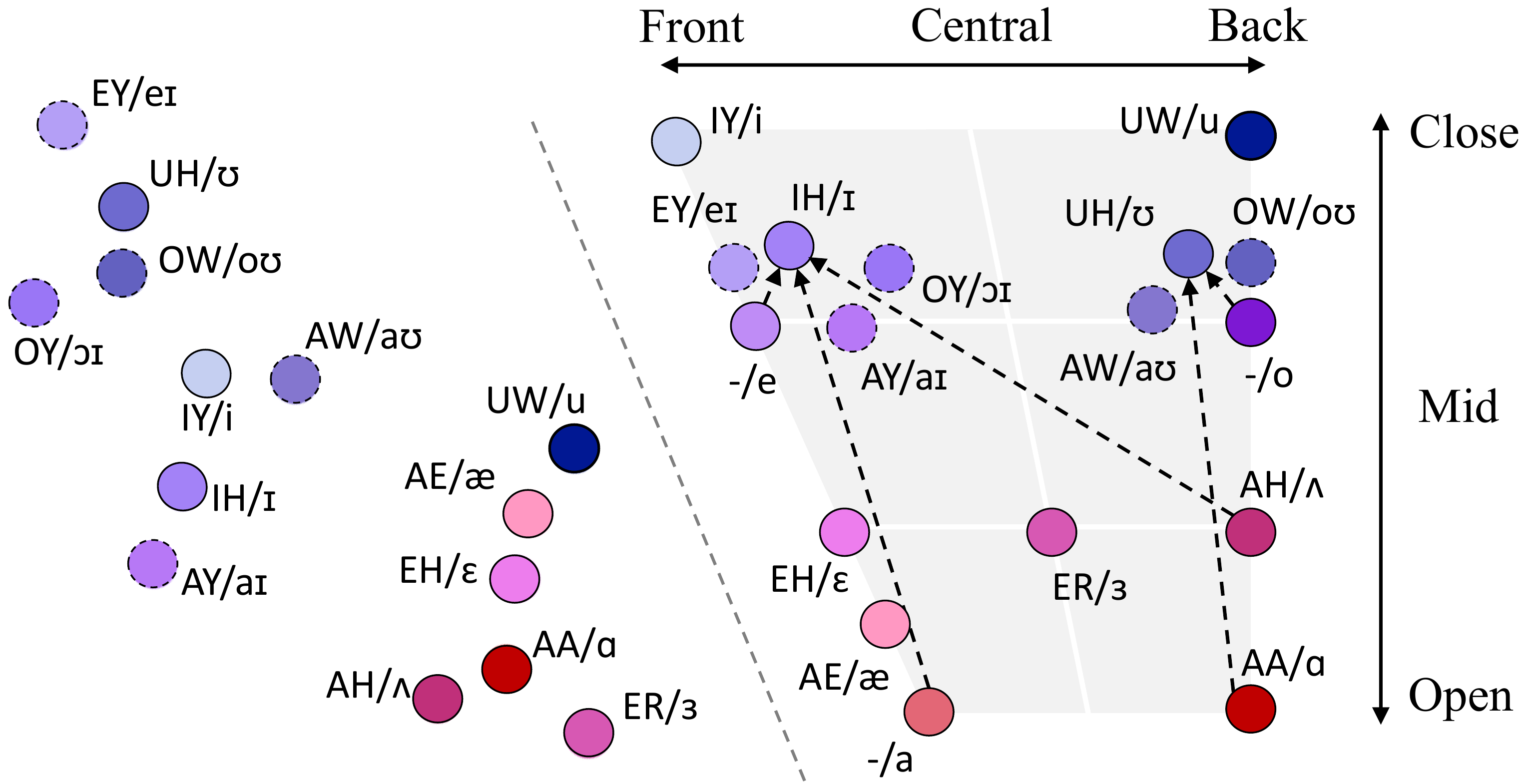}}
\vspace{-7pt}
\caption{\small Comparison of the learned representation and phoneme units.
The left part is the t-SNE visualization of vowel representations  from our codebook trained with 22 hours of unpaired speech and 20 minutes of paired data.
The right part is the IPA vowel chart defined by linguists with the corresponding ARPABET.}
\vspace{-10pt}
\label{fig:vowel}
\end{figure}
\section{EXPERIMENTS}
Experiments were performed on LJSpeech~\cite{ljspeech17} which consists of 13,100 audio clips ($\approx$ 24 hours) of a single female speaker. 
We followed the prior work~\cite{ren2019almost} to randomly choose the development/test set with 300 audio clips in each set and a different amount of paired data (5/15/20 minutes) from the remaining data.
For the unselected data, we discarded the transcription and treated them as unpaired speech.
We followed the previous work of TTS~\cite{wang2017tacotron} to extract spectrogram with the window size of 50~ms and the hop size of 12.5~ms.
For the linguistic units, CMU phoneme set~\cite{zhang2003identifying} is used for grapheme to phoneme conversion~\cite{g2pE2019}.

The encoder is composed of a 7-layer convolution network with 512 kernels for each layer followed by 2-layer LSTMs with 512 cells.
Tacotron 2~\cite{shen2018natural} is adopted as the decoder and an additional CBHG module as in Tacotron~\cite{wang2017tacotron} is used to predict spectrogram from mel spectrogram. 
Griffin-Lim algorithm~\cite{griffin1984signal} is used to convert spectrogram to waveform and adapting a vocoder is left as future work.
The codebook contains 40 entries to match the size of the phoneme set and each entry is a randomly initialized vector of 64 dimensions.
To meet the request of CTC objective described in Eq.~(\ref{eq:ctc_prob}), one entry of the codebook is used as the blank token and we simply omit the corresponding embedding vector when performing temporal segmentation.

To objectively evaluate the effectiveness of our proposed representation learning framework, we also conducted experiments of our proposed framework without learning representation by removing the codebook (which we referred to \textit{ours without codebook} throughout our experiments).
In this setting, the encoder directly predicts the probability over phoneme set for each frame.
The phoneme index sequence is obtained by choosing the most probable phone of each frame and perform temporal segmentation.
The decoder takes a sequence of phoneme index (instead of embedding) and maintains its own embedding table as a normal TTS~\cite{wang2017tacotron}.
For speech reconstruction, temporal segmentation is performed on pseudo one-hot categorical distribution (with ST gradient estimation) outputted by the encoder.
This can be regarded as a special case of speech chain with ST-estimator~\cite{tjandra2019end} where the ASR is a CTC network with temporal segmentation and the unpaired text is not utilized.

\subsection{Vowel Representation Parallel to IPA Vowel Chart}
\label{subsec:repr}
To interpret the quantized speech representation, we visualized the codebook learned by SeqRQ-AE in Fig.~\ref{fig:vowel} and compared it against the IPA vowel chart~\cite{international1999handbook} defined by linguists.
On the right-hand side, we colored the IPA vowel chart according to the position of the tongue. 
Blue means the tongue is close to the roof of the mouth, red indicates the opposite. 
The color is darker when the highest point of the tongue is positioned relatively back in the mouth.
A t-SNE~\cite{maaten2008visualizing} visualization of the learned vowel embedding was shown on the left-hand side and we colored each vowel with respect to its color assigned in IPA vowel chart. 
We can observe that the color distributions of these two sides were quite similar. 
The \textit{front} and \textit{close} vowels (colored bright and blue) in the IPA vowel chart grouped on the upper left region of our visualization. 
On the other hand, most of the \textit{back} and \textit{open} vowels (colored dark and red) located in the lower right region. 
The fact that the relationship between representations learned by SeqRQ-AE matched the relationship between phonemes defined by experts indicates that SeqRQ-AE is capable of learning meaningful phonetic embedding. 
In the following sections, we further applied the learned representation to perform ASR and TTS tasks.

\vspace{-7pt}
\subsection{Speech recognition}
\label{subsec:asr}
\begin{table}[t]
\small
\centering
\begin{threeparttable}

\caption{\small Phoneme error rate (\%) on different amount of paired data.}

\begin{tabular}{l| c| c| c| c}
\toprule
\multicolumn{1}{l|}{Method}  & 20 min & 15 min & 10 min & 5 min\\ \hline 
Baseline          & 29.4 & 33.2 & 41.2 & 55.7\\    
Ren et al.~\cite{ren2019almost}\textsuperscript{\textdagger} & 11.7 & -    & 64.2 & -   \\ \hline \hline   
Ours & & & & \\ \hline
- w/o codebook & 29.9 & 33.2 & 41.4 & 56.2  \\ 
- w/~ codebook  & 25.5 & 29.0 & 35.2 & 49.3 \\

\bottomrule
\end{tabular}
\vspace{-2pt}
\begin{tablenotes}
\item{\textdagger} \small{\text{Used unpaired text besides unpaired speech.}}
\end{tablenotes}
\vspace{-7pt}
\label{table:per}
\end{threeparttable}
\end{table}

To perform speech recognition, we selected the most possible phoneme sequence according to the distance between each encoder output and phoneme embedding in the codebook (see Eq.~(\ref{eq:prob}) and (\ref{eq:seq_prob})) with beam search and trimmed the repeated phonemes.
Table~\ref{table:per} shows the phoneme error rate (PER) of the speech recognition task. 
The baseline is an ASR model (which is not required to reconstruct the input speech nor to learn any representation) having the same architecture as our encoder with an additional projection layer to predict probability over phoneme set.

For all the amounts of paired data considered, our method defeated the baseline ASR.
We also discovered that without representation learning, our framework performed similarly to the baseline ASR.
Although the model proposed by Ren et al.~\cite{ren2019almost} had a better performance than our method with 20 minutes paired data, in the 10-minute setting, our method outperformed all other models by a significant gap. 
We suspected the phoneme representation learned across the encoder and the decoder is the key to our success, since in all other cases (the model proposed by Ren et al, our model w/o codebook, and the baseline) such embedding does not exist.
With all the pieces of evidence mentioned above, we conclude that the representations learned from unpaired speech with SeqRQ-AE can significantly improve ASR when the amount of paired data is extremely rare.

\vspace{-7pt}
\subsection{Text-to-speech synthesis}
\label{subsec:tts}
In TTS experiment, we randomly sampled 50 sentences from the test set to conduct the Mean Opinion Score (MOS) test and listed the result in Table~\ref{table:mos}.
50 subjects were asked to rate the given audio according to naturalness and each utterance at least received 5 ratings.
For all the models evaluated in Table~\ref{table:mos}, we initialized 16 (out of 64) dimensions of the codebook (or input embedding for pure TTS model) with pre-defined phoneme attributes~\cite{Tong:256326} at the beginning of the training process to generate speech with higher quality. 
The differential spectral loss~\cite{shechtman2019sequence} was adopted to boost the performance of TTS model.
We also compared our method against Speech Chain~\cite{tjandra2017listening}, a dual learning framework for ASR and TTS where the two modules do not share representation.
The ASR and TTS were trained on paired data and pseudo paired data derived from self-labeled unpaired data.

\begin{figure}[t]
\centerline{\includegraphics[width=8.5cm]{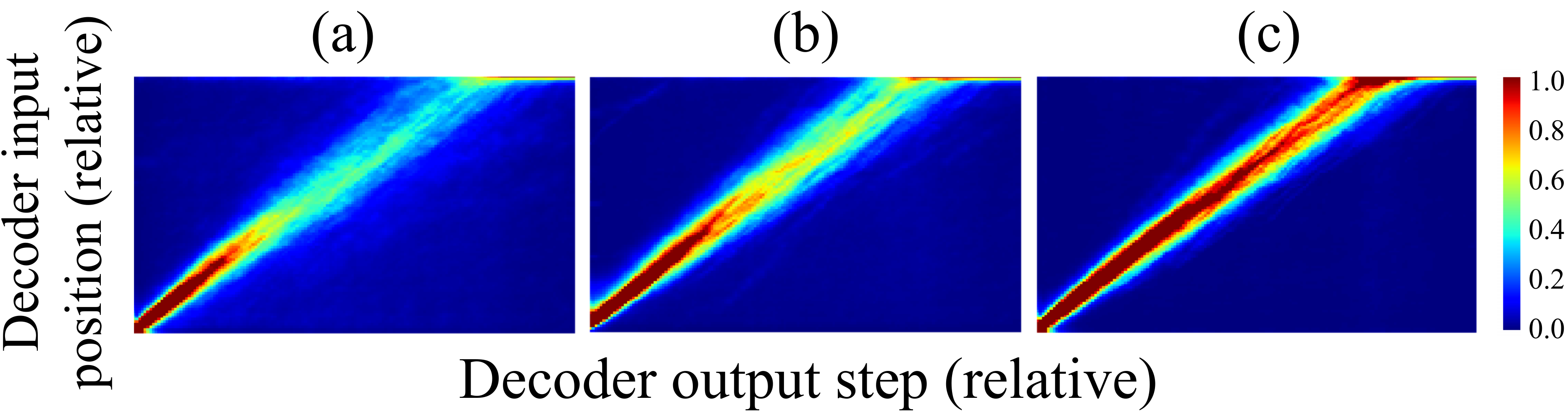}}
\caption{\small The overall alignment of three different models with 20 minutes of paired data. (a) Speech Chain~\cite{tjandra2017listening} (w/o unpaired text) (b) our method w/o codebook (c) our method.
The completeness of the diagonal of each method aligned to the number of mistakes it made.
Since the total number of the decoder input positions and the total number of the decoding steps for each alignment varies, we resized them to a fixed number before taking the average.}
\vspace{-10pt}
\label{fig:align}
\end{figure}

The result showed our method outperformed Speech Chain (without text-to-text cycle) when there were only 20 minutes of paired data available. 
This is because Speech Chain can only generate short utterance and failed to utter intelligible speech for longer sentences.
To analyze the ability to complete utterances, we took a look into the generated alignments which were well known to be strongly correlated to the robustness of TTS model. 
In Fig.~\ref{fig:align}, we showed the overall alignments by normalizing and averaging all the alignments found by different models in the test set. 
The more prominent and more complete the diagonal, the better the capability of the model to complete an utterance.
The result in Fig.~\ref{fig:align} showed our model (part (c)) was more robust for generating intelligible speech for long input text than models without codebook (part (b)) while Speech Chain (part (c)) could hardly finish most of the sentences.
To further verify our hypothesis, 100 generated outputs in the test set for each model were checked by humans to see whether there were mistakes (word repeating, word skipping or word error) without considering the naturalness.
We found the number of mistakes made by our 10min/20min(no codebook)/20min model, 71/51/10 respectively, matched the results of MOS test (row(\romannum{6})(\romannum{4})(\romannum{5}) in Table~\ref{table:mos}) and alignment robustness  (part(a)(b)(c) in Fig.~\ref{fig:align}).
All these results demonstrated the fact that representations learned from unpaired data benefit TTS when access to annotated data is limited.
Samples drawn from our model are provided on the webpage\footnote{\url{https://ttaoretw.github.io/SeqRQ-AE/demo.html}}.
\begin{table}[t]
\small
\centering
\begin{threeparttable}

\caption{\small Mean Opinion Score (MOS) ratings with 95\% confidence intervals for naturalness.}

\begin{tabular}{c| l| c| c c}
\toprule
& \multicolumn{1}{l|}{Method}  & Paired Data   & \multicolumn{1}{c}{MOS}  \\ \hline
(\romannum{1}) & Ground truth                & -             &   4.81$\pm$0.026   \\
(\romannum{2}) & Fully-supervised            &  23 hr~~      &   3.55$\pm$0.038   \\ \hline
(\romannum{3}) & Speech Chain~\cite{tjandra2017listening}\textsuperscript{\textsection}               & \multirow{3}{*}{20 min} &   1.92$\pm$0.038   \\ 
(\romannum{4}) & Ours w/o codebook           &               &   2.33$\pm$0.040   \\ 
(\romannum{5}) & Ours                        &               &   2.62$\pm$0.037   \\ \hline
(\romannum{6}) & Ours                        & 10 min        &   1.69$\pm$0.034   \\

\bottomrule
\end{tabular}
\vspace{-2pt}
\begin{tablenotes}
\item{\textsection} \small{\text{Trained without using unpaired text.}}
\end{tablenotes}
\label{table:mos}
\end{threeparttable}
\end{table}

\section{Conclusion}
\label{sec:conclusion}

In this work, we introduce \QuatizeName~AutoEncoder (SeqRQ-AE), a novel framework for learning quantized speech representation corresponded to the underlying linguistic units. 
The experiments showed that the learned representation contains phonetic information aligned with the phoneme relationship defined by linguists and is also excessively helpful for ASR and TTS with very limited paired data.
In the future, we aim to leverage unpaired text to our framework and pursue fully unsupervised speech recognition and synthesis.

\newpage
\vfill\pagebreak
\bibliographystyle{IEEEbib}
{\bibliography{strings,refs}}
\end{document}